\documentclass[11pt]{article}

\usepackage[margin=1in]{geometry}
\usepackage[T1]{fontenc}
\usepackage[utf8]{inputenc}
\usepackage{lmodern}
\usepackage{microtype}
\usepackage{setspace}
\usepackage{amsmath, amssymb, amsthm, mathtools}
\usepackage{booktabs}
\usepackage{multirow}
\usepackage{graphicx}
\usepackage{xcolor}
\usepackage{hyperref}
\usepackage[nameinlink,capitalize]{cleveref}
\usepackage{enumitem}
\usepackage{listings}
\usepackage{algorithm}
\usepackage{algpseudocode}
\usepackage[section]{placeins}

\hypersetup{
  colorlinks=true,
  linkcolor=blue!55!black,
  citecolor=blue!55!black,
  urlcolor=blue!55!black,
  pdftitle={Behavioral Consistency in Machine Learning},
  pdfauthor={Lan Wu, Bruce Changlong Xu}
}

\lstdefinelanguage{yaml}{
  keywords={true,false,null,True,False,Null,yes,no,on,off},
  sensitive=true,
  comment=[l]{\#},
  morestring=[b]',
  morestring=[b]"
}

\theoremstyle{plain}

\newtheorem{proposition}{Proposition}
\newtheorem{lemma}{Lemma}

\theoremstyle{definition}
\newtheorem{definition}{Definition}

\theoremstyle{remark}

\newcommand{\ktrain}{K_{\mathrm{train}}}
\newcommand{\kinf}{K_{\mathrm{inf}}}
\newcommand{\contract}{\mathcal{C}}
\newcommand{\R}{\mathbb{R}}
\newcommand{\E}{\mathbb{E}}
\newcommand{\KL}{\mathrm{KL}}
\newcommand{\ece}{\mathrm{ECE}}
\newcommand{\TV}{\mathrm{TV}}
\newcommand{\argmax}{\mathop{\mathrm{arg\,max}}}
\newcommand{\policy}{\pi}
\newcommand{\polktrain}{\policy^{\ktrain}}
\newcommand{\polkinf}{\policy^{\kinf}}
\newcommand{\spanop}{\mathrm{span}}

\title{\textbf{Behavioral Consistency in Machine Learning}}
\author{%
  \textbf{Lan Wu}\\
  University of California, Berkeley
  \and
  \textbf{Bruce Changlong Xu}\\
  Stanford University
}
\date{}

\begin{document}
\maketitle

\begin{abstract}
A trained model is usually described by its weights, as if the weights fixed its behavior. In practice the same weights are run by different programs. Training uses high-precision accumulation and fixed tensor shapes, while serving uses low-precision arithmetic, fused attention kernels, a paged key-value cache, and dynamic batching. These programs produce different next-token distributions from identical weights. We make this dependence explicit by indexing a policy by the kernel that executes it, and we propose \emph{kernel contracts}, which state how far a serving kernel may drift from a reference kernel on each slice of traffic and what the serving system should do when a limit is exceeded. We prove non-asymptotic bounds that link the logit difference between two kernels to total-variation and KL divergence, to the change in expected reward, to the bias of policy gradients computed from rollouts generated by another kernel, and to whether the greedy token can change. We test the bounds and the contracts in GPU experiments on \texttt{Qwen2.5-7B-Instruct} with Hugging Face and vLLM. The bounds hold on every prompt we measured. Lower precision, a different serving engine, and a different batch size each change the generated token on some prompts, and these changes concentrate on prompts where the model is nearly undecided between two tokens. A kernel can therefore look fine on aggregate metrics while behaving differently on exactly these prompts. Checking a contract separately on each slice of traffic catches these cases.
\end{abstract}

\section{Introduction}
\label{sec:introduction}

We usually describe a trained language model by its weights $\theta$ and write its policy as $\policy_\theta$. This notation hides a detail that matters in practice. The same weights are executed by different programs at different stages of a model's life. During training, a framework such as PyTorch evaluates $\policy_\theta$ with high-precision accumulators, fixed batch shapes, and operators chosen so that gradients are stable. During serving, and during rollout generation for reinforcement learning, an inference engine evaluates the same weights with low-precision arithmetic or quantized copies of the weights \cite{micikevicius2022fp8,frantar2023gptq,lin2023awq}, a paged key-value cache \cite{kwon2023vllm}, and attention kernels whose order of summation depends on the current batch \cite{dao2022flashattention,dao2023flashattention2,shah2024flashattention3}. Floating-point addition is not associative \cite{goldberg1991every,higham2002accuracy}, so these programs return different logits for the same input. One checkpoint therefore behaves like several slightly different models, depending on where it runs. We refer to the question of how much these models may differ as behavioral consistency.

On average the differences are small, which is why aggregate benchmark scores rarely move when a model is moved to a new serving stack. They are not small everywhere. When the two most likely tokens are nearly tied, a small change in the logits can change which token is generated, and in autoregressive decoding that change carries into every later step.

Reinforcement-learning post-training is especially exposed. PPO \cite{schulman2017ppo,christiano2017deep,ouyang2022instructgpt} and GRPO \cite{shao2024deepseekmath} pipelines often generate rollouts with a fast serving engine such as vLLM \cite{kwon2023vllm} and compute gradients with a separate training engine. The update treats both engines as the same policy, which amounts to setting the importance ratio between them, $w(x,y) = \polktrain_\theta(y\mid x)/\polkinf_\theta(y\mid x)$, to one. When the engines disagree, this biases the gradient \cite{precup2000eligibility,degris2012off,munos2016safe,espeholt2018impala} and uses up part of the PPO clipping range before any parameter has changed. The documented sensitivity of deep reinforcement learning to implementation details \cite{henderson2018deep,engstrom2020implementation,andrychowicz2021what} is consistent with this mechanism.

Similar problems appear outside reinforcement learning. Calibration \cite{guo2017calibration} measured under one kernel need not hold under another, a quantized model can answer differently on prompts that sit close to a decision such as whether to refuse, and a request served by a continuous-batching engine \cite{kwon2023vllm} can receive a different greedy continuation depending on what else is in its batch. In each case the checkpoint and the serving engine are versioned separately, and nothing states how much they are allowed to disagree.

We propose writing that statement down as a \emph{kernel contract}, following the design-by-contract idea from software engineering \cite{meyer1992applying}. A contract lists numerical, statistical, runtime, and observability conditions on a pair of kernels, the slices of traffic on which each condition must hold, and the action the serving system takes when a condition fails. The actions are to promote the new kernel, log the violation, route affected requests to a safer configuration, or fall back to the reference kernel. Requiring bit-for-bit equality between training and serving would give up the speed that low-precision kernels and batching provide. A contract instead states which differences are acceptable and on which inputs, so that they can be checked before and after deployment.

The paper makes four contributions.
\begin{enumerate}[leftmargin=1.8em,itemsep=0.25em]
  \item We index the policy by the kernel that executes it and define contracts over pairs of kernels, with per-slice thresholds, exceedance budgets, and three severity levels (\Cref{sec:preliminaries,sec:motivating-cases,sec:contract-framework}).
  \item We prove non-asymptotic bounds from the logit difference $\varepsilon(x) = z^{\kinf}_\theta(x) - z^{\ktrain}_\theta(x)$ to total-variation distance, KL divergence, and expected reward, and a margin condition under which the greedy token cannot change (\Cref{sec:mismatch}).
  \item We bound the policy-gradient bias caused by rollouts from a different kernel, give the bias that remains after clipping the importance ratio, and relate both to the PPO clipping range (\Cref{sec:rl-application}). \Cref{sec:enforcement} describes how a contract gates promotion and routing.
  \item We measure nine Hugging Face and vLLM configurations of \texttt{Qwen2.5-7B-Instruct} \cite{yang2024qwen25} on a single GPU over 500 TruthfulQA prompts \cite{lin2022truthfulqa}. We separate the effects of attention fusion, precision, engine, and batch size, find no violations of the bounds in 7{,}000 prompt-level comparisons, and show that greedy-token changes concentrate on low-margin prompts that aggregate checks miss (\Cref{sec:experiments}).
\end{enumerate}

\section{Kernel-Indexed Policies and Measures of Drift}
\label{sec:preliminaries}

A \emph{kernel} $K$ is the program that maps weights $\theta \in \R^d$ and a context $x \in \mathcal{X}$ to next-token logits $z^K_\theta(x) \in \R^V$ over a vocabulary $\mathcal{V}$ of size $V$, and that also computes gradients when it is used for training. At temperature $T > 0$ it induces the next-token distribution
\begin{equation}
\policy^K_\theta(v \mid x) \;=\; \frac{\exp(z^K_\theta(x)_v / T)}{\sum_{u \in \mathcal{V}} \exp(z^K_\theta(x)_u / T)}, \qquad v \in \mathcal{V}.
\label{eq:kernel-policy}
\end{equation}
Two kernels that load the same weights generally return different logits, because floating-point addition is not associative and low-precision formats round intermediate values \cite{goldberg1991every,higham2002accuracy}. We therefore keep the kernel as an explicit superscript.

We write $\ktrain$ for a reference kernel used in training or evaluation, for example PyTorch with FP32 or BF16 accumulation and fixed batch shapes, and $\kinf$ for a candidate serving or rollout kernel, for example vLLM \cite{kwon2023vllm} with PagedAttention, FlashAttention \cite{dao2022flashattention,dao2023flashattention2,shah2024flashattention3}, FP16, BF16, or FP8 arithmetic \cite{micikevicius2022fp8}, and continuous batching. Inputs are drawn from an evaluation distribution $D_{\mathrm{eval}}$. A \emph{slice} $S \subseteq \mathcal{X}$ is a subset of inputs that we want to check separately, such as prompts on which the model is nearly undecided, long prompts, or prompts on medical or legal topics. We write $\Sigma = \{S_1, \dots, S_m\}$ for the set of slices.

\subsection{Measuring disagreement between two kernels}

Fix $\theta$ and an input $x$. We compare $\polktrain_\theta$ and $\polkinf_\theta$ in six ways.

\paragraph{Logit difference.} Let $\varepsilon(x) = z^{\kinf}_\theta(x) - z^{\ktrain}_\theta(x) \in \R^V$. We report its largest absolute entry $\|\varepsilon(x)\|_\infty$, its Euclidean norm $\|\varepsilon(x)\|_2$, and its range over the vocabulary,
\begin{equation}
\spanop\bigl(\varepsilon(x)\bigr) \;=\; \max_{v \in \mathcal{V}} \varepsilon_v(x) \;-\; \min_{u \in \mathcal{V}} \varepsilon_u(x) \;\le\; 2\,\|\varepsilon(x)\|_\infty.
\label{eq:logit-span}
\end{equation}
Adding the same constant to every logit leaves the softmax unchanged, so the range is the part of $\varepsilon(x)$ that can change probabilities.

\paragraph{Distribution difference.} We use the total-variation distance and the KL divergence,
\begin{equation}
\Delta_{\TV}(x) \;=\; \tfrac{1}{2}\sum_{v\in\mathcal{V}} \bigl| \polktrain_\theta(v\mid x) - \polkinf_\theta(v\mid x) \bigr|,
\qquad
\Delta_{\KL}(x) \;=\; \KL\bigl(\polktrain_\theta(\cdot\mid x) \,\Vert\, \polkinf_\theta(\cdot\mid x)\bigr).
\label{eq:tv-kl}
\end{equation}

\paragraph{Importance ratio.} When a token $y$ is sampled from $\polkinf_\theta$ and scored under $\polktrain_\theta$, the importance ratio is
\begin{equation}
w(x,y) \;=\; \frac{\polktrain_\theta(y \mid x)}{\polkinf_\theta(y \mid x)}.
\label{eq:importance-ratio}
\end{equation}
Under $y \sim \polkinf_\theta(\cdot\mid x)$ it satisfies $\E[w] = 1$ and $\E[|w - 1|] = 2\,\Delta_{\TV}(x)$, and its expected value of $w \log w$ equals the KL divergence above:
\begin{equation}
\E_{y \sim \polkinf_\theta(\cdot\mid x)}\bigl[w(x,y) \log w(x,y)\bigr] \;=\; \sum_{y \in \mathcal{V}} \polktrain_\theta(y\mid x) \log \frac{\polktrain_\theta(y\mid x)}{\polkinf_\theta(y\mid x)} \;=\; \Delta_{\KL}(x).
\label{eq:wlogw-kl}
\end{equation}
A KL limit can therefore be checked from rollout tokens alone.

\paragraph{Rank agreement and margin.} Let $\mathrm{TopK}_k(z)$ be the $k$ highest-scoring tokens under logits $z$. The top-$k$ overlap is
\begin{equation}
A_{\mathrm{top}k}(x) \;=\; \frac{1}{k}\,\bigl|\mathrm{TopK}_k(z^{\ktrain}_\theta(x)) \cap \mathrm{TopK}_k(z^{\kinf}_\theta(x))\bigr|.
\label{eq:topk}
\end{equation}
For $k = 1$ it records whether greedy decoding picks the same token under both kernels. How easily that token changes depends on the reference margin
\begin{equation}
\gamma(x) \;=\; z^{\ktrain}_{\theta,(1)}(x) \;-\; z^{\ktrain}_{\theta,(2)}(x),
\label{eq:margin}
\end{equation}
the gap between the largest and second-largest reference logits. A small margin means the two leading tokens are nearly tied.

\paragraph{Calibration and confidence.} On a slice $S$, the calibration gap $\Delta_{\ece}(S) = |\ece(\polktrain_\theta; S) - \ece(\polkinf_\theta; S)|$ compares expected calibration error \cite{guo2017calibration}, and $\Delta_{\mathrm{conf}}(x) = |\max_v \polktrain_\theta(v\mid x) - \max_v \polkinf_\theta(v\mid x)|$ is the change in top-1 probability.

\paragraph{Reward and runtime.} For a reward $r : \mathcal{X} \times \mathcal{Y} \to [-R, R]$, the change in expected reward is
\begin{equation}
\Delta_r(x) \;=\; \E_{y \sim \polktrain_\theta(\cdot\mid x)} [r(x,y)] \;-\; \E_{y \sim \polkinf_\theta(\cdot\mid x)} [r(x,y)].
\label{eq:reward-drift}
\end{equation}
At runtime we also record latency percentiles, throughput, peak memory, and whether repeated runs reproduce the same logits.

\subsection{Contracts and severity levels}

A kernel contract checks some of these quantities on chosen slices. A clause's metric is either a per-input quantity, such as $\E_{y}|\log w(x,y)|$, or a statistic of a slice, such as the 95th percentile of $\|\varepsilon(x)\|_\infty$ or the mean of $\Delta_{\KL}(x)$. A \emph{hard} clause must hold on every slice it covers. A \emph{soft} clause on a per-input metric may fail on at most a fraction $\rho$ of the inputs in each slice.

\begin{definition}[Contract satisfaction]
\label{def:contracted-convergence}
Let $\contract$ have hard clauses $\mathcal{H}$ and soft clauses $\mathcal{S}_{\mathrm{soft}}$, where clause $c$ has metric $m_c$, threshold $\tau_c$, slices $\Sigma_c \subseteq \Sigma$, and, if soft, budget $\rho_c \in (0,1]$. The pair $(\ktrain, \kinf)$ \emph{satisfies} $\contract$ on $D_{\mathrm{eval}}$ if
\begin{equation}
m_h(S) \le \tau_h \;\; (\forall\, h \in \mathcal{H},\, S \in \Sigma_h)
\quad\text{and}\quad
\Pr_{x \sim D_{\mathrm{eval}} \mid S}\bigl[ m_s(x) > \tau_s \bigr] \le \rho_s \;\; (\forall\, s \in \mathcal{S}_{\mathrm{soft}},\, S \in \Sigma_s),
\label{eq:contracted-convergence}
\end{equation}
where $m_h(S)$ is the slice statistic of a hard clause (for a per-input metric, its maximum over $S$).
\end{definition}

Each clause carries a severity level that decides what happens when it fails:
\begin{itemize}[leftmargin=1.8em,itemsep=0.15em]
  \item \textbf{L1} violations are logged and do not block traffic.
  \item \textbf{L2} violations freeze promotion or route the affected slice to a higher-precision kernel or to an importance-corrected update.
  \item \textbf{L3} violations roll back to the reference kernel.
\end{itemize}

\section{Four Ways Kernel Differences Change Model Behavior}
\label{sec:motivating-cases}

The four cases below arise in different parts of the stack. In each, a pipeline treats $\policy_\theta$ as one function when it is in fact evaluated by several kernels.

\subsection{Rollouts from a serving engine bias the policy gradient}
\label{sec:case-rlhf}

Policy-gradient methods such as TRPO \cite{schulman2015trpo}, PPO \cite{schulman2017ppo,ouyang2022instructgpt}, and GRPO \cite{shao2024deepseekmath} sample from the current policy and update the weights along
\begin{equation}
\nabla_\theta J(\theta) \;=\; \E_{x \sim D, \, y \sim \policy_\theta(\cdot\mid x)}\bigl[ \nabla_\theta \log \policy_\theta(y \mid x)\, A(x,y) \bigr].
\end{equation}
Many post-training pipelines split this work between two engines. Rollouts come from an inference engine $\kinf$ such as vLLM \cite{kwon2023vllm}, chosen for throughput, while log-probabilities and gradients come from a training engine $\ktrain$ such as PyTorch with FSDP. Even at the first step of an update, when both engines hold the same weights, their forward passes disagree. An estimator that ignores this computes
\begin{equation}
\hat{g}(\theta) \;=\; \E_{x \sim D,\, y \sim \polkinf_\theta(\cdot\mid x)}\bigl[ \nabla_\theta \log \polktrain_\theta(y\mid x)\, A(x,y) \bigr],
\end{equation}
which differs from the on-policy gradient $g^\star(\theta)$ whenever $w(x,y) \neq 1$. If the pipeline also uses the rollout engine's log-probabilities as the denominator of the PPO ratio, the kernel difference moves that ratio away from one before the weights have changed. On \texttt{Qwen2.5-7B-Instruct} (\Cref{sec:experiments}), sampling from vLLM in BF16 and scoring in Hugging Face FP32 gives $|w(x,y) - 1| > 0.05$ for $11.5\%$ of sampled tokens. A deviation of $0.05$ is a quarter of a typical PPO clipping range of $0.2$.

\subsection{Lower precision moves the logits by an amount set by the mantissa}

Models evaluated with FP32 accumulation are often served in BF16, FP16, or FP8 \cite{micikevicius2022fp8,frantar2023gptq,lin2023awq}. BF16 keeps 8 exponent bits and 7 stored mantissa bits, so its unit roundoff is $2^{-8} \approx 3.9 \times 10^{-3}$. FP16 keeps 5 exponent bits and 10 mantissa bits, with unit roundoff $2^{-11} \approx 4.9 \times 10^{-4}$ \cite{goldberg1991every,higham2002accuracy}. When values stay within FP16's range, BF16 rounding errors are therefore about eight times larger. The FP8 E4M3 format keeps 3 mantissa bits, with unit roundoff $2^{-4}$. In our measurements, rounding every linear-layer weight of a 28-layer model to E4M3 changes the logits by about one unit in $\ell_\infty$ norm. A token that leads the next candidate by several logits keeps its rank under such a change. A token that leads by a fraction of a logit often does not.

\subsection{Fused and unfused attention agree in FP32 but not in BF16}

A textbook (eager) attention implementation computes $QK^\top/\sqrt{d_k}$, rounds this matrix and its softmax to the working precision, and then multiplies by $V$. Fused implementations such as PyTorch's scaled dot-product attention (SDPA) and FlashAttention \cite{dao2022flashattention,dao2023flashattention2,shah2024flashattention3} keep intermediate values in FP32 registers and do not round the attention matrix between the two products. In FP32 the two implementations agree to within $10^{-4}$ in every logit. In BF16 the eager version differs from fused attention by a mean $\|\varepsilon\|_\infty$ of $2.08$ and changes the greedy token on $8.0\%$ of prompts (\Cref{sec:experiments}).

\subsection{Batch composition changes the logits}

Continuous-batching engines such as vLLM \cite{kwon2023vllm} group incoming requests into batches whose size changes with the traffic. Matrix-multiplication and attention kernels can choose tile sizes and reduction orders according to the number of tokens in a batch, so the logits for a fixed prompt can depend on what else is in its batch. On a single GPU with fixed weights and BF16 precision, running a prompt alone ($B=1$) instead of in a batch of 50 changes the logits by a mean $\|\varepsilon\|_\infty$ of $0.28$ and changes the greedy token on $1.0\%$ of prompts (\Cref{tab:decomp}). Fleets that mix accelerator types or engine versions add further variation. An evaluation that runs one prompt at a time and a production service that batches requests are therefore measuring slightly different models.

\section{Kernel Contracts}
\label{sec:contract-framework}

A kernel contract states, for a pair of kernels $(\ktrain, \kinf)$, which differences are acceptable and what to do when they are not. Following the design-by-contract idea \cite{meyer1992applying}, we write it as
\begin{equation}
\contract \;=\; (\mathcal{N}, \mathcal{S}, \mathcal{R}, \mathcal{O}, \Pi).
\end{equation}
The first four entries are sets of numerical, statistical, runtime, and observability clauses, and $\Pi$ is an escalation policy that maps violations to actions. Each clause is a record
\begin{equation}
\gamma \;=\; \bigl(\,\mathrm{id},\; m,\; \tau,\; \Sigma_\gamma,\; \rho,\; \lambda,\; \pi\,\bigr)
\end{equation}
with an identifier, a metric $m$ from \Cref{sec:preliminaries}, a threshold $\tau$, the slices $\Sigma_\gamma \subseteq \Sigma$ on which it is checked, an exceedance budget $\rho \in [0,1]$ ($\rho = 0$ for a hard clause), a severity level $\lambda \in \{\mathrm{L1, L2, L3}\}$, and a remediation action $\pi$.

\subsection{Four kinds of clause}

\paragraph{Numerical clauses ($\mathcal{N}$).} These limit differences in logits and probabilities before any token is sampled. Examples are quantiles of $\|\varepsilon(x)\|_\infty$, $\spanop(\varepsilon(x))$, $\|\varepsilon(x)\|_2$, $\Delta_{\TV}(x)$, and $\Delta_{\KL}(x)$, and, for reinforcement learning, limits on $\E_{y\sim\pi_b}[|\log w(x,y)|]$ and $\E_{y\sim\pi_b}[w(x,y)\log w(x,y)]$. They catch changes caused by precision casts, weight quantization, and operator fusion.

\paragraph{Statistical clauses ($\mathcal{S}$).} These limit changes that users or a training loop would see after sampling. They cover greedy agreement $A_{\mathrm{top1}}$, top-$k$ overlap $A_{\mathrm{top}k}$, calibration gap $\Delta_{\ece}$, confidence change $\Delta_{\mathrm{conf}}$, and changes in reward or refusal rate $\Delta_r$. They are usually checked on slices that matter most, such as low-margin or safety-relevant prompts.

\paragraph{Runtime clauses ($\mathcal{R}$).} These limit latency percentiles, throughput, and peak memory, including the key-value cache. Without them, a candidate kernel could pass the numerical clauses by disabling batching or by falling back to a slow reference path.

\paragraph{Observability clauses ($\mathcal{O}$).} These require each kernel to record what is needed to check the other clauses later. The records include build identifiers (engine and CUDA versions, attention backend, quantization scheme, hardware), request shapes, and a replay test that reruns the same inputs under the same batch schedule and compares the logits. Without these records, a violation cannot be traced to its cause.

\subsection{Hard and soft clauses, escalation, and versioning}

Some quantities should never exceed their limit, such as the mean KL divergence between kernels or the difference between two replays of the same request. These are hard clauses ($\rho = 0$). Others, such as the per-prompt deviation of the importance ratio or the latency of individual requests, are expected to exceed a threshold occasionally. A soft clause allows this on at most a fraction $\rho$ of the inputs in each slice.

The escalation policy takes the most severe violation across all clauses and slices and selects an action,
\begin{equation}
\Pi : \{\mathrm{pass, L1, L2, L3}\} \;\longrightarrow\; \{\,\text{promote},\; \text{log},\; \text{guard},\; \text{fallback}\,\}.
\end{equation}
If every clause passes, the candidate kernel is promoted. An L1 violation is logged. An L2 violation triggers a guarded route, for example sending low-margin prompts to an FP16 kernel or turning on importance-ratio correction for reinforcement-learning rollouts. An L3 violation blocks the candidate and falls back to the reference kernel.

The contract is versioned together with the weights and the serving build. If the hashes of all three are recorded at each promotion, a later regression can be traced to a change in the weights, the kernel, or the contract thresholds.

\section{Logit Differences Bound Changes in Distribution, Reward, and Greedy Tokens}
\label{sec:mismatch}

A contract is easiest to check at the level of logits, but what we care about is behavior. This section connects the two. We first split the logit difference by its source, then bound the resulting change in the next-token distribution, in expected reward, and in the greedy token.

\subsection{Splitting the logit difference by source}

For a prompt $x$, let $\varepsilon(x) = z^{\kinf}_\theta(x) - z^{\ktrain}_\theta(x)$. Moving from $\ktrain$ to $\kinf$ usually changes several things at once. If we pass through intermediate kernels that change one thing at a time, the logit difference splits into a sum of terms, one per change:
\begin{equation}
\varepsilon(x) \;=\; \varepsilon_{\mathrm{approx}}(x) \;+\; \varepsilon_{\mathrm{quant}}(x) \;+\; \varepsilon_{\mathrm{route}}(x) \;+\; \varepsilon_{\mathrm{sched}}(x).
\label{eq:decomposition}
\end{equation}
Here $\varepsilon_{\mathrm{approx}}$ comes from replacing one operator implementation with an algebraically equivalent one, for example eager attention with fused attention \cite{dao2022flashattention,dao2023flashattention2,shah2024flashattention3}. The term $\varepsilon_{\mathrm{quant}}$ comes from changing the number format, for example FP32 to BF16, FP16, or FP8 weights \cite{micikevicius2022fp8}. The term $\varepsilon_{\mathrm{route}}$ comes from changing the engine at fixed precision, for example from Hugging Face Transformers to vLLM with PagedAttention and fused normalization kernels \cite{kwon2023vllm}. The term $\varepsilon_{\mathrm{sched}}$ comes from changing the batch in which the prompt is run, for example from a batch of one to a batch of 50. The terms depend on the order in which the changes are applied, so in \Cref{sec:experiments} we also measure each change on its own. By the triangle inequality, $\|\varepsilon(x)\|_\infty$ is at most the sum of the four terms' $\ell_\infty$ norms.

\subsection{From logits to distributions, rewards, and greedy tokens}

The following bounds hold for every $\varepsilon(x) \in \R^V$ and every vocabulary size $V \ge 2$. They have no higher-order remainder terms.

\begin{lemma}[Logit difference bounds the change in distribution]
\label{lem:logit-to-dist}
Let $p = \mathrm{softmax}(z/T)$ and $q = \mathrm{softmax}((z+\varepsilon)/T)$ over a vocabulary $\mathcal{V}$ at temperature $T > 0$, and let $\spanop(\varepsilon) = \max_{v} \varepsilon_v - \min_{u} \varepsilon_u \le 2\,\|\varepsilon\|_\infty$. Then
\begin{align}
\Delta_{\TV}(p, q) &\;\le\; \tanh\left(\frac{\spanop(\varepsilon)}{4T}\right) \;\le\; \frac{\spanop(\varepsilon)}{4T} \;\le\; \frac{\lVert \varepsilon \rVert_\infty}{2T}, \label{eq:lem-tv}\\
\KL(p \,\Vert\, q) &\;\le\; \frac{\spanop(\varepsilon)^2}{8T^2} \;\le\; \frac{\lVert \varepsilon \rVert_\infty^2}{2T^2}, \label{eq:lem-kl}\\
\exp\left(-\frac{\spanop(\varepsilon)}{T}\right) &\;\le\; \frac{p(v)}{q(v)} \;\le\; \exp\left(\frac{\spanop(\varepsilon)}{T}\right) \qquad \text{for all } v \in \mathcal{V}. \label{eq:lem-ratio}
\end{align}
\end{lemma}

The proof (\Cref{app:proof-logit-to-dist}) interpolates between the two logit vectors and bounds the first two cumulants of $\varepsilon$ along the path with Popoviciu's inequality. The $\tanh$ bound is attained by a two-token distribution in which one token's logit rises by $\spanop(\varepsilon)$ relative to the other and that token has probability $1/(1 + e^{\spanop(\varepsilon)/(2T)})$. The linear constants $1/(4T)$ and $1/(2T)$ are therefore the best possible as $\varepsilon \to 0$.

The next lemma states when a logit difference cannot change the greedy token.

\begin{lemma}[Greedy token is stable when the margin exceeds the logit range]
\label{lem:margin-flip}
Let $\gamma(x) > 0$ be the reference margin defined in \Cref{eq:margin}. If
\begin{equation}
\spanop\bigl(\varepsilon(x)\bigr) \;<\; \gamma(x),
\label{eq:margin-condition}
\end{equation}
which holds in particular when $\|\varepsilon(x)\|_\infty < \gamma(x)/2$, then $\argmax_{v} z^{\kinf}_\theta(x)_v = \argmax_{v} z^{\ktrain}_\theta(x)_v$ and $A_{\mathrm{top1}}(x) = 1$.
\end{lemma}

The proof is in \Cref{app:proof-margin-flip}. The lemma explains how a kernel can agree with the reference on more than $98\%$ of prompts and still change the greedy token on more than $20\%$ of the prompts in one slice. Prompts with a small margin need only a small logit difference to swap their two leading tokens.

Finally, total-variation distance bounds the change in any bounded reward.

\begin{proposition}[Change in expected reward]
\label{prop:reward-bound}
For every input $x$ and every reward $r : \mathcal{X} \times \mathcal{Y} \to [-R, R]$,
\begin{equation}
\bigl|\Delta_r(x)\bigr| \;\le\; 2R \cdot \Delta_{\TV}(x) \;\le\; \frac{R}{T}\,\lVert \varepsilon(x) \rVert_\infty.
\label{eq:prop-reward}
\end{equation}
\end{proposition}

Together, \Cref{lem:logit-to-dist} and \Cref{prop:reward-bound} mean that a limit on the logit difference, which can be measured directly on the kernels, implies a limit on the change in expected reward. If $\lVert \varepsilon(x) \rVert_\infty \le \tau$, then $|\Delta_r(x)| \le R\tau/T$.

\section{Rollouts from a Different Kernel Bias the Policy Gradient}
\label{sec:rl-application}

When rollouts are sampled with $\kinf$ and gradients are computed with $\ktrain$, the logit difference between the two kernels becomes a bias in the gradient estimate. This section bounds that bias, gives the bias that remains after clipping the importance ratio, and relates the result to PPO's clipping range.

\subsection{Setup and assumptions}
\label{sec:rl-assumptions}

Let $x$ be a prompt together with the tokens generated so far, and let $y \in \mathcal{V}$ be the next token. The behavior policy that generates rollouts is $\pi_b(\cdot \mid x) = \polkinf_\theta(\cdot \mid x)$, the target policy that is differentiated is $\pi_t(\cdot \mid x) = \polktrain_\theta(\cdot \mid x)$, and the per-token importance ratio is $w(x,y) = \pi_t(y\mid x)/\pi_b(y\mid x)$ \cite{precup2000eligibility,degris2012off,munos2016safe,espeholt2018impala}. We work at the level of a single token and assume:
\begin{enumerate}[label=(A\arabic*),leftmargin=2.2em,itemsep=0.15em]
  \item $\pi_b(y\mid x) > 0$ whenever $\pi_t(y\mid x) > 0$. This holds for any softmax policy with finite logits.
  \item $|A(x,y)| \leq A_{\max}$ for all $(x,y)$. This holds when rewards or group-relative advantages are normalized or clipped.
  \item $\lVert \nabla_\theta \log \pi_t(y\mid x) \rVert_2 \leq G$ for all $(x,y)$.
\end{enumerate}

\subsection{Bias of the uncorrected estimator}

Rewritten as an expectation over rollouts from $\pi_b$, the on-policy gradient of $\pi_t$ is
\begin{equation}
g^\star(\theta) \;=\; \E_{x \sim D,\, y \sim \pi_b(\cdot\mid x)}\bigl[ w(x,y)\, \nabla_\theta \log \pi_t(y\mid x)\, A(x,y) \bigr].
\end{equation}
A pipeline that samples with one engine and differentiates with another, without correcting for the difference, computes instead
\begin{equation}
\hat g(\theta) \;=\; \E_{x \sim D,\, y \sim \pi_b(\cdot\mid x)}\bigl[\nabla_\theta \log \pi_t(y\mid x)\, A(x,y) \bigr].
\end{equation}

\begin{proposition}[Policy-gradient bias from a kernel mismatch]
\label{prop:rl-bias}
Under \textnormal{(A1)--(A3)}, the bias $b(\theta) = \hat g(\theta) - g^\star(\theta)$ satisfies
\begin{equation}
\lVert b(\theta) \rVert_2 \;\leq\; A_{\max} G \, \E_{x \sim D}\bigl[\lVert \pi_t(\cdot\mid x) - \pi_b(\cdot\mid x) \rVert_1\bigr] \;=\; 2 A_{\max} G \, \E_x\bigl[ \Delta_{\TV}(x) \bigr] \;\le\; \frac{A_{\max} G}{T}\,\E_x\bigl[\lVert \varepsilon(x) \rVert_\infty\bigr].
\label{eq:prop-rl-bias}
\end{equation}
By Pinsker's inequality, $\Delta_{\TV}(x) \le \sqrt{\Delta_{\KL}(x)/2}$, and with \Cref{eq:wlogw-kl} this gives
\begin{align}
\lVert b(\theta) \rVert_2 &\;\le\; \sqrt{2}\,A_{\max}\,G \cdot \E_x\left[\sqrt{\E_{y\sim\pi_b}\bigl[w(x,y)\log w(x,y)\bigr]}\right] \nonumber\\
&\;\le\; A_{\max}\,G\,\sqrt{2\,\E_{x,\,y\sim\pi_b}\bigl[w(x,y)\log w(x,y)\bigr]}.
\label{eq:prop-rl-pinsker}
\end{align}
\end{proposition}

The proof is in \Cref{app:proof-rl-bias}. Both the mean total-variation distance and the mean of $w \log w$, which can be estimated from rollout tokens, bound the size of the bias.

\subsection{Clipping the importance ratio reduces the bias}
\label{sec:rl-correction}

Multiplying each term by the exact ratio $w(x,y)$ removes the bias but can increase variance when a rare token has a large ratio. A common compromise clips the ratio to $[1-c, 1+c]$ for some clip width $c \ge 0$:
\begin{equation}
\tilde g_c(\theta) \;=\; \E_{x \sim D,\,y \sim \pi_b(\cdot\mid x)}\bigl[ \mathrm{clip}\bigl(w(x,y),\, 1-c,\, 1+c\bigr)\, \nabla_\theta \log \pi_t(y\mid x)\, A(x,y) \bigr].
\label{eq:clipped-pg}
\end{equation}
Setting $c = 0$ recovers the uncorrected estimator $\hat g$.

\begin{proposition}[Bias after clipping the importance ratio]
\label{prop:clipped-bias}
For any $c \ge 0$, the weight that \Cref{eq:clipped-pg} places on token $y$ is $\pi_b(y\mid x)\,\mathrm{clip}(w(x,y), 1-c, 1+c) = \mathrm{clip}(\pi_t(y\mid x),\allowbreak (1-c)\pi_b(y\mid x),\allowbreak (1+c)\pi_b(y\mid x))$. Under \textnormal{(A1)--(A3)}, the bias $b_c(\theta) = \tilde{g}_c(\theta) - g^\star(\theta)$ satisfies
\begin{equation}
\lVert b_c(\theta) \rVert_2 \;\le\; A_{\max}\, G \cdot \E_{x \sim D}\left[\sum_{y \in \mathcal{V}} \Bigl(\bigl|\pi_t(y\mid x) - \pi_b(y\mid x)\bigr| - c\,\pi_b(y\mid x)\Bigr)_{+}\right],
\label{eq:clipped-bias-bound}
\end{equation}
where $(u)_+ = \max(u, 0)$. The right-hand side equals $2A_{\max}G\,\E_x[\Delta_{\TV}(x)]$ at $c = 0$, is non-increasing and convex in $c$, and is zero once $c \ge \sup_{x,y} |w(x,y) - 1|$.
\end{proposition}

The proof is in \Cref{app:proof-clipped-bias}. The same reasoning applies to PPO. With clipping parameter $\epsilon_{\mathrm{clip}}$, usually between $0.1$ and $0.2$, PPO stops the gradient of a token once the ratio $\pi_t(y\mid x)/\pi_b(y\mid x)$ leaves $[1-\epsilon_{\mathrm{clip}}, 1+\epsilon_{\mathrm{clip}}]$ \cite{schulman2017ppo}. If the denominator comes from the rollout engine, then at the start of an update, before $\theta$ has moved, this ratio already equals the kernel ratio $w(x,y)$. Tokens with $|w(x,y) - 1| > \epsilon_{\mathrm{clip}}$ are clipped before any learning has happened, and tokens with a smaller but nonzero deviation have less room to move in one direction than in the other.

\subsection{Contract clauses for reinforcement-learning pipelines}

\Cref{prop:rl-bias,prop:clipped-bias} suggest four clauses for pipelines that sample and train with different engines:
\begin{itemize}[leftmargin=1.8em,itemsep=0.15em]
  \item \textbf{N-RLHF-1} (soft): on every slice $S$, the fraction of prompts with $\E_{y \sim \pi_b}[|\log w(x,y)|] > \tau_w$ is at most $\rho_w$.
  \item \textbf{N-RLHF-2} (hard): on every slice $S$, the mean of $\E_{y \sim \pi_b}[w(x,y) \log w(x,y)]$, which equals the mean of $\Delta_{\KL}(x)$, is at most $\epsilon_{\KL}$.
  \item \textbf{S-RLHF-1}: the greedy change rate $1 - \E_{x \in S}[A_{\mathrm{top1}}(x)]$ is at most $\tau_{\mathrm{flip}}$, and the top-5 overlap is at least $\tau_{\mathrm{top5}}$.
  \item \textbf{R-RLHF-1}: rollout latency and throughput stay within their limits at the allotted key-value-cache memory.
\end{itemize}

\section{Using a Contract to Gate Promotion and Routing}
\label{sec:enforcement}

The four kinds of clause need different conditions to be checked. Numerical and statistical clauses can be checked offline on a fixed set of prompts, runtime clauses need realistic load, and observability clauses must hold throughout. We therefore check a contract at several stages before promotion and keep checking it afterwards.

\subsection{Four stages before full promotion}

Every candidate change, whether new weights, a new serving engine, a new quantization scheme, or new hardware, passes through four stages.
\begin{enumerate}[leftmargin=1.8em,itemsep=0.2em]
  \item \textbf{Offline check.} All numerical, statistical, and observability clauses, and single-machine runtime clauses, are evaluated on a fixed set of prompts divided into the slices $\Sigma$. An L3 violation stops the candidate. An L2 violation either blocks promotion or sends the candidate to a guarded configuration.
  \item \textbf{Shadow traffic.} Live requests are copied to the candidate kernel, and its outputs are recorded but not returned to users. This checks runtime and numerical clauses under real batch sizes and prompt lengths, which a fixed offline batch does not exercise.
  \item \textbf{Canary.} The candidate serves a small fraction $\alpha \in (0, \alpha_{\max}]$ of traffic with the escalation policy active. Observability records allow any flagged request to be replayed against $\ktrain$.
  \item \textbf{Full promotion.} If the candidate stays within the contract for a period $W$ at fraction $\alpha_{\max}$, it serves all traffic. A sample of requests continues to be checked against the reference kernel.
\end{enumerate}

\subsection{Routing during serving}

During canary and production serving, we summarize compliance over a recent window $[t-w, t]$ with a health score
\begin{equation}
h_t \;=\; 1 - \max_{c \in \mathcal{H} \cup \mathcal{S}_{\mathrm{soft}}} \, \mathrm{normalize}\bigl( m_c(\cdot)_{[t-w, t]},\, \tau_c \bigr).
\end{equation}
Here $\mathrm{normalize}$ divides the measured value (or, for a soft clause, the measured exceedance rate) by its limit, oriented so that values above one mean the clause is violated. The score is therefore non-negative exactly when every clause holds. With thresholds $\eta_2 > \eta_3$, the router serves requests on the candidate kernel when $h_t \ge \eta_2$, uses the guarded route for flagged slices when $\eta_3 \le h_t < \eta_2$, and falls back to the reference kernel when $h_t < \eta_3$. \Cref{alg:promotion-loop} summarizes the procedure.

\begin{algorithm}[t]
\caption{Contract-based promotion and per-slice routing}
\label{alg:promotion-loop}
\begin{algorithmic}[1]
\Require Contract $\contract = (\mathcal{N},\mathcal{S},\mathcal{R},\mathcal{O},\Pi)$, candidate $(\theta', \kinf')$, reference $(\theta, \ktrain)$, slices $\Sigma$
\State \textbf{Offline check:} evaluate every clause on every slice $S \in \Sigma$
\If{some clause is violated at level L3}
  \State \Return fall back to $(\theta, \ktrain)$
\ElsIf{some clause is violated at level L2 on slice $S$}
  \State Set up a guarded route for $S$ (for example, send $x \in S$ to an FP16 kernel or clip $w$)
\EndIf
\State \textbf{Shadow and canary:} copy traffic for a period $W_s$, then serve a fraction $\alpha \in (0, \alpha_{\max}]$
\For{each request $x$ during the canary period $W$}
  \State Route $x$ according to its slice and the current health score $h_t$
  \State Periodically replay sampled requests against $\ktrain$ and update $h_t$
\EndFor
\If{$h_t \geq \eta_2$ throughout $W$}
  \State Promote $(\theta', \kinf')$ and record $(\mathrm{hash}(\theta'), \mathrm{hash}(\kinf'), \mathrm{hash}(\contract))$
\Else
  \State Roll back to $(\theta, \ktrain)$ and record the violated clauses and slices
\EndIf
\end{algorithmic}
\end{algorithm}

\section{Experiments}
\label{sec:experiments}

\subsection{Setup}

We run all experiments on a single GPU with \texttt{Qwen2.5-7B-Instruct} \cite{yang2024qwen25}, which has 28 layers and produces logits over $V = 152{,}064$ token indices. The inputs are the first 500 questions of the TruthfulQA validation set \cite{lin2022truthfulqa}, formatted with the model's chat template. Their median length is 40 tokens, the 95th percentile is 52, and the longest is 78. For each prompt and kernel we record the full logit vector for the first generated token in double precision and use temperature $T = 1$. Every kernel loads the same weights and receives the same token sequences, so any difference between two kernels comes from the kernels themselves.

The reference kernel $\ktrain$ is Hugging Face Transformers in FP32 with PyTorch's fused scaled-dot-product attention (\texttt{hf-fp32}). We also run FP32 with eager attention (\texttt{hf-fp32-eager}) to measure the effect of fusion alone. The seven candidate kernels vary four things. The engine is either Hugging Face Transformers (with PyTorch 2.11) or vLLM 0.10.2 (with PyTorch 2.8). The number format is FP16, BF16, or BF16 with FP8 weights, which we simulate by rounding every linear-layer weight to E4M3 with a per-output-channel scale and computing in BF16. The attention implementation is fused or eager. For vLLM, prompts are run either one at a time ($B=1$) or in batches of 50. We also record latency, throughput, and peak memory, and we rerun the first 25 prompts to test whether each kernel reproduces its own logits. All nine kernels reproduced them exactly.

\begin{table}[ht]
\centering
\caption{Difference between each candidate kernel and the reference \texttt{hf-fp32} on 500 TruthfulQA prompts. $\E[w\log w]$ is the mean KL divergence $\Delta_{\KL}(x)$, computed over the full vocabulary. ``Greedy change'' is the fraction of prompts whose most likely token differs from the reference. Latency is the 95th percentile per request; for batched vLLM runs it is wall-clock time per prompt averaged over the batch. Throughput counts prompt tokens processed per second.}
\label{tab:main-kernels}
\vspace{0.35em}
\resizebox{\linewidth}{!}{%
\begin{tabular}{lcccccccc}
\toprule
& \multicolumn{2}{c}{Logit difference $\|\varepsilon(x)\|_\infty$} & \multicolumn{2}{c}{Total variation $\Delta_{\TV}(x)$} & Mean KL & Greedy & Top-5 & Latency / \\
\cmidrule(lr){2-3} \cmidrule(lr){4-5}
Candidate kernel $\kinf$ & Mean & p95 (max) & Mean & p95 (max) & $\E[w\log w]$ & change & overlap & throughput \\
\midrule
\texttt{hf-fp16}               & $0.0330$ & $0.0530$ ($0.126$) & $0.00121$ & $0.00380$ ($0.0113$) & $1.14 \times 10^{-5}$ & $0.6\%$ ($3/500$)  & $99.84\%$ & $433.0\text{ ms}$ / $998\text{ tok/s}$ \\
\texttt{vllm-fp16} ($B=50$)    & $0.0310$ & $0.0468$ ($0.171$) & $0.00125$ & $0.00405$ ($0.0094$) & $1.17 \times 10^{-5}$ & $0.6\%$ ($3/500$)  & $99.92\%$ & $1.5\text{ ms}$ / $27{,}212\text{ tok/s}$ \\
\texttt{vllm-bf16} ($B=50$)    & $0.2644$ & $0.4165$ ($0.668$) & $0.00935$ & $0.02994$ ($0.0526$) & $6.25 \times 10^{-4}$ & $1.4\%$ ($7/500$)  & $98.20\%$ & $1.6\text{ ms}$ / $26{,}452\text{ tok/s}$ \\
\texttt{vllm-bf16-b1} ($B=1$)  & $0.2636$ & $0.4182$ ($0.848$) & $0.01005$ & $0.03208$ ($0.0900$) & $7.69 \times 10^{-4}$ & $1.2\%$ ($6/500$)  & $98.36\%$ & $19.1\text{ ms}$ / $2{,}988\text{ tok/s}$ \\
\texttt{hf-bf16}               & $0.2571$ & $0.3917$ ($0.797$) & $0.01014$ & $0.03309$ ($0.0918$) & $8.00 \times 10^{-4}$ & $1.8\%$ ($9/500$)  & $98.60\%$ & $433.6\text{ ms}$ / $989\text{ tok/s}$ \\
\texttt{hf-fp8-e4m3}           & $0.9743$ & $1.4941$ ($2.342$) & $0.03580$ & $0.11977$ ($0.2423$) & $1.04 \times 10^{-2}$ & $4.4\%$ ($22/500$) & $95.52\%$ & $12.5\text{ ms}$ / $3{,}325\text{ tok/s}$ \\
\texttt{hf-bf16-eager}         & $2.0671$ & $4.8759$ ($14.35$) & $0.07129$ & $0.26110$ ($0.9313$) & $6.73 \times 10^{-2}$ & $7.4\%$ ($37/500$) & $91.24\%$ & $17.7\text{ ms}$ / $2{,}756\text{ tok/s}$ \\
\bottomrule
\end{tabular}%
}
\end{table}

\subsection{Each source of drift is measurable on its own}

\begin{figure}[t]
\centering
\includegraphics[width=\linewidth]{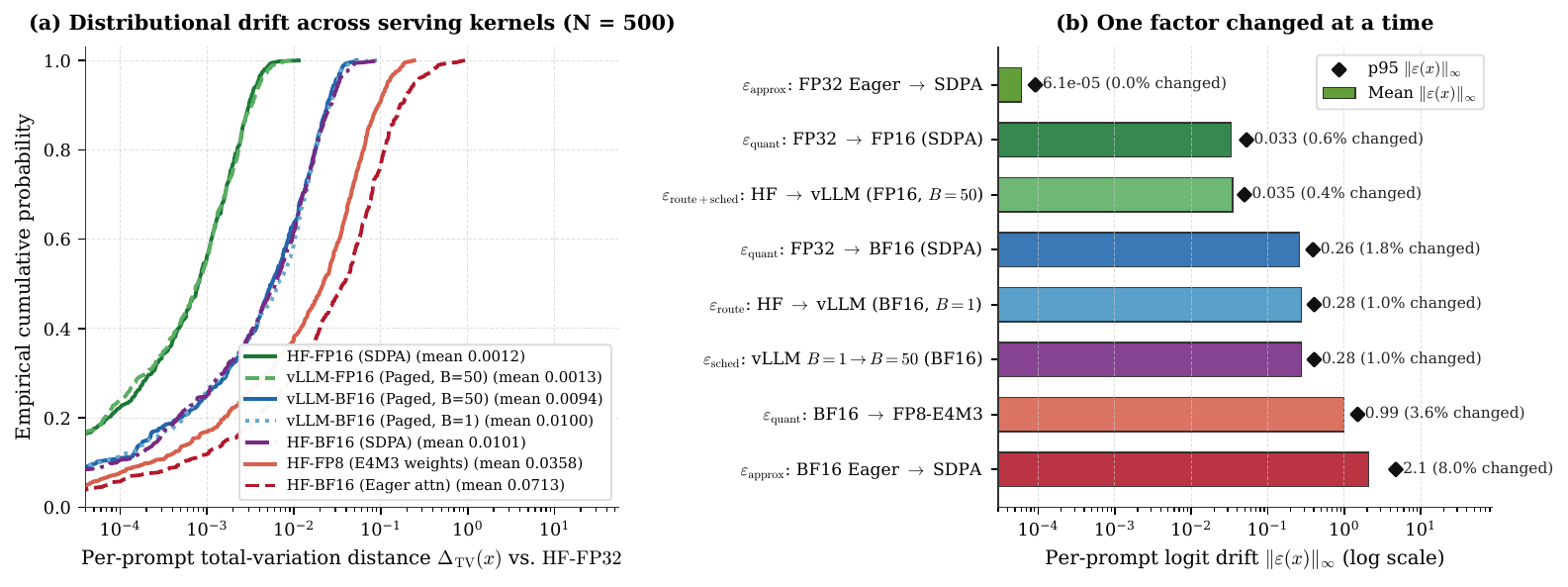}
\caption{Kernel drift on \texttt{Qwen2.5-7B-Instruct} over 500 TruthfulQA prompts. \textbf{(a)} Empirical distribution of the per-prompt total-variation distance from the \texttt{hf-fp32} reference. \textbf{(b)} Mean and 95th-percentile logit difference $\|\varepsilon(x)\|_\infty$ when one factor is changed at a time (\Cref{tab:decomp}), with the fraction of prompts whose greedy token changes.}
\label{fig:drift-overview}
\end{figure}

\Cref{tab:main-kernels} and \Cref{fig:drift-overview}(a) compare each candidate with the FP32 reference. FP16 stays much closer to FP32 than BF16 does. Both FP16 kernels have a mean $\|\varepsilon\|_\infty$ of $0.031$ to $0.033$, a mean total-variation distance of about $0.0012$, and a mean KL divergence of about $1.2 \times 10^{-5}$, and they change the greedy token on $0.6\%$ of prompts (3 of 500). The three BF16 kernels have a mean $\|\varepsilon\|_\infty$ of $0.257$ to $0.264$ and a mean total-variation distance of $0.0094$ to $0.0101$, both about eight times the FP16 values. Their mean KL divergence is 50 to 70 times larger, and they change the greedy token on $1.2\%$ to $1.8\%$ of prompts. The factor of eight matches the ratio of the two formats' unit roundoffs, $2^{-8}/2^{-11}$. FP8 weights increase the drift further, to a mean $\|\varepsilon\|_\infty$ of $0.974$ (95th percentile $1.49$) and a mean total-variation distance of $0.036$, and change the greedy token on $4.4\%$ of prompts.

\Cref{tab:decomp} and \Cref{fig:drift-overview}(b) change one factor at a time. Attention fusion makes almost no difference in FP32, where eager and fused attention differ by a mean $\|\varepsilon\|_\infty$ of $6.1 \times 10^{-5}$ and never change the greedy token. In BF16 the same switch gives a mean $\|\varepsilon\|_\infty$ of $2.08$ and changes the greedy token on $8.0\%$ of prompts, because eager attention rounds the attention matrix to BF16 between the two matrix products. Changing the engine at fixed BF16 precision and $B=1$, from Hugging Face to vLLM, gives a mean total-variation distance of $0.0116$, slightly more than the $0.0101$ caused by casting FP32 to BF16 within Hugging Face. With vLLM batching at $B=50$ the engine change gives $0.0103$. Within vLLM in BF16, changing only the batch size from 1 to 50 gives a mean $\|\varepsilon\|_\infty$ of $0.278$ and a mean total-variation distance of $0.0110$, and changes the greedy token on $1.0\%$ of prompts.

\begin{table}[t]
\centering
\caption{Changing one factor at a time. Each row compares two kernels that differ in the listed factor only, except the two rows marked route + sched, which change the engine and also run vLLM with batches of 50. The first column refers to the terms of \Cref{eq:decomposition}.}
\label{tab:decomp}
\vspace{0.35em}
\resizebox{\linewidth}{!}{%
\begin{tabular}{llcccccc}
\toprule
Term and factor changed & Kernel pair & Mean $\|\varepsilon\|_\infty$ & p95 $\|\varepsilon\|_\infty$ & Mean $\|\varepsilon\|_2$ & Mean $\Delta_{\TV}$ & $\E[w\log w]$ & Greedy change \\
\midrule
$\varepsilon_{\mathrm{approx}}$: attention fusion, FP32 & \texttt{hf-fp32-eager} $\to$ \texttt{hf-fp32}             & $6.11 \times 10^{-5}$ & $9.21 \times 10^{-5}$ & $0.0038$ & $3.04 \times 10^{-6}$ & $6.95 \times 10^{-11}$ & $0.0\%$ ($0/500$) \\
$\varepsilon_{\mathrm{quant}}$: FP32 $\to$ FP16          & \texttt{hf-fp32} $\to$ \texttt{hf-fp16}                   & $0.0330$              & $0.0530$              & $2.252$  & $0.00121$             & $1.14 \times 10^{-5}$  & $0.6\%$ ($3/500$) \\
$\varepsilon_{\mathrm{route+sched}}$: engine, FP16       & \texttt{hf-fp16} $\to$ \texttt{vllm-fp16}                 & $0.0347$              & $0.0499$              & $2.456$  & $0.00149$             & $1.72 \times 10^{-5}$  & $0.4\%$ ($2/500$) \\
$\varepsilon_{\mathrm{quant}}$: FP32 $\to$ BF16          & \texttt{hf-fp32} $\to$ \texttt{hf-bf16}                   & $0.2571$              & $0.3917$              & $17.69$  & $0.01014$             & $8.00 \times 10^{-4}$  & $1.8\%$ ($9/500$) \\
$\varepsilon_{\mathrm{route}}$: engine, BF16, $B=1$      & \texttt{hf-bf16} $\to$ \texttt{vllm-bf16-b1}              & $0.2774$              & $0.4062$              & $19.66$  & $0.01156$             & $1.09 \times 10^{-3}$  & $1.0\%$ ($5/500$) \\
$\varepsilon_{\mathrm{route+sched}}$: engine, BF16       & \texttt{hf-bf16} $\to$ \texttt{vllm-bf16}                 & $0.2702$              & $0.3829$              & $19.10$  & $0.01030$             & $9.64 \times 10^{-4}$  & $1.2\%$ ($6/500$) \\
$\varepsilon_{\mathrm{sched}}$: batch $1 \to 50$, BF16   & \texttt{vllm-bf16-b1} $\to$ \texttt{vllm-bf16}            & $0.2775$              & $0.4062$              & $20.07$  & $0.01099$             & $9.48 \times 10^{-4}$  & $1.0\%$ ($5/500$) \\
$\varepsilon_{\mathrm{quant}}$: BF16 $\to$ FP8 weights   & \texttt{hf-bf16} $\to$ \texttt{hf-fp8-e4m3}               & $0.9911$              & $1.5000$              & $74.40$  & $0.03647$             & $1.04 \times 10^{-2}$  & $3.6\%$ ($18/500$) \\
$\varepsilon_{\mathrm{approx}}$: attention fusion, BF16 & \texttt{hf-bf16-eager} $\to$ \texttt{hf-bf16}             & $2.0815$              & $4.7266$              & $134.7$  & $0.07184$             & $8.13 \times 10^{-2}$  & $8.0\%$ ($40/500$) \\
\bottomrule
\end{tabular}%
}
\end{table}

\subsection{The bounds hold on every prompt, and greedy changes follow the margin}

\begin{figure}[t]
\centering
\includegraphics[width=\linewidth]{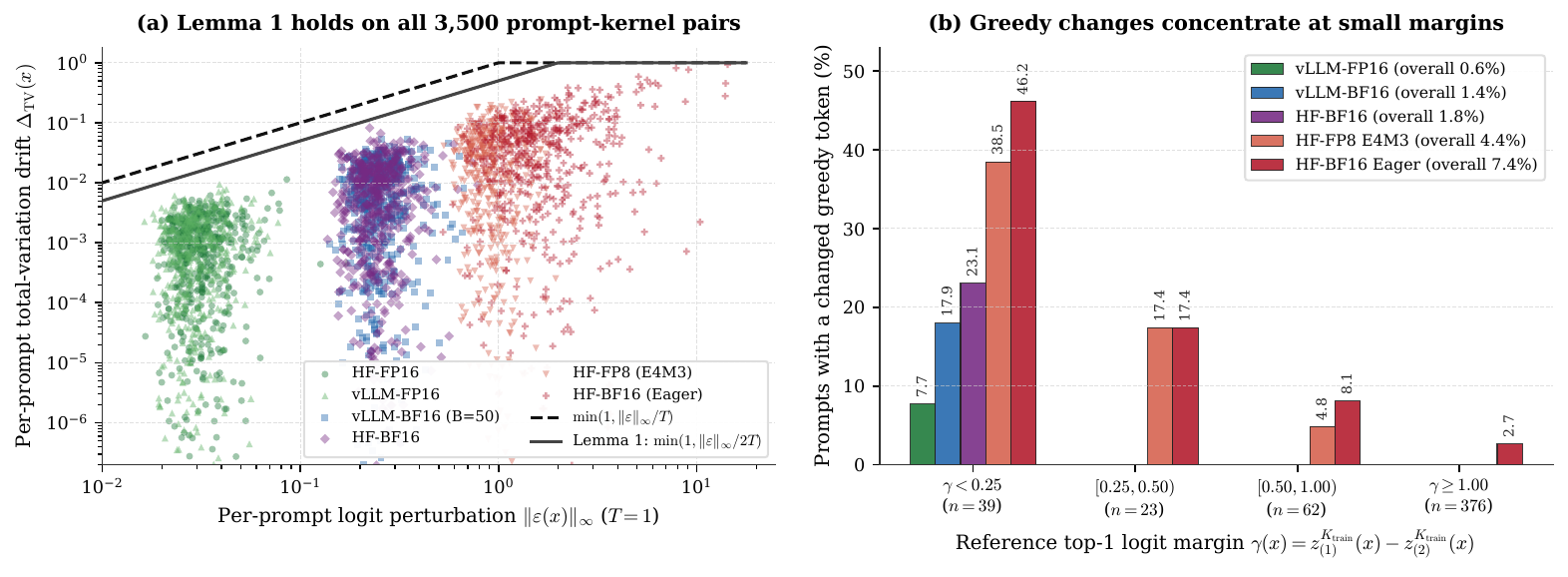}
\caption{\textbf{(a)} Per-prompt total-variation distance against $\|\varepsilon(x)\|_\infty$ for the seven candidate kernels (3{,}500 points), with the bounds $\min(1, \|\varepsilon\|_\infty/T)$ and $\min(1, \|\varepsilon\|_\infty/2T)$. No point lies above either line. \textbf{(b)} Fraction of prompts whose greedy token changes, grouped by the reference margin $\gamma(x)$.}
\label{fig:bounds-margins}
\end{figure}

We checked \Cref{lem:logit-to-dist} on all 14 kernel pairs in \Cref{tab:main-kernels,tab:decomp}, which gives $14 \times 500 = 7{,}000$ prompt-level comparisons. No comparison violates the total-variation bound $\Delta_{\TV}(x) \le \spanop(\varepsilon(x))/(4T) \le \|\varepsilon(x)\|_\infty/(2T)$ or the KL bound $\Delta_{\KL}(x) \le \spanop(\varepsilon(x))^2/(8T^2)$ (\Cref{fig:bounds-margins}(a)). The bound is loose in practice. The mean total-variation distance is about $7\%$ of the mean of $\|\varepsilon(x)\|_\infty/(2T)$, because most of the probability sits on a few tokens while $\|\varepsilon(x)\|_\infty$ is the largest change over all 152{,}064 tokens.

\Cref{fig:bounds-margins}(b) groups prompts by the reference margin $\gamma(x)$, which has a median of $2.31$, a 10th percentile of $0.35$, and a 90th percentile of $7.16$. Thirty-nine prompts have $\gamma(x) < 0.25$ and 23 have $0.25 \le \gamma(x) < 0.50$. For all five FP16 and BF16 kernels, every greedy change occurred on one of the 39 prompts with $\gamma(x) < 0.25$. On those prompts the change rate was $7.7\%$ for both FP16 kernels, $15.4\%$ for vLLM-BF16 at $B=1$, $17.9\%$ for vLLM-BF16 at $B=50$, and $23.1\%$ for HF-BF16. On the other 461 prompts it was zero. With FP8 weights, whose 95th-percentile $\|\varepsilon(x)\|_\infty$ is $1.49$, changes also occur for $0.25 \le \gamma(x) < 0.50$ ($17.4\%$) and $0.50 \le \gamma(x) < 1.00$ ($4.8\%$), but none of the 376 prompts with $\gamma(x) \ge 1.00$ changed. BF16 eager attention, which has the largest logit differences, also changes $2.7\%$ of these high-margin prompts. This pattern is consistent with \Cref{lem:margin-flip}. Kernels with larger logit differences change the greedy token on prompts with larger margins.

\subsection{Serving-engine rollouts use part of the PPO clipping range before training starts}

For each kernel we compute, over the full vocabulary, the probability that a token sampled from $\kinf$ has $|w(x,y) - 1|$ above $0.05$, $0.10$, and $0.20$. We also draw 20 tokens per prompt (10{,}000 in total) to show the spread of sampled ratios. \Cref{tab:rl-bias} reports both. With BF16 sampling and FP32 scoring, $11.5\%$ to $12.8\%$ of sampled tokens deviate from one by more than $0.05$, and $2.9\%$ to $3.9\%$ by more than $0.10$, before the weights have changed. With FP8 weights, $34.7\%$ of tokens deviate by more than $0.05$ and $10.7\%$ by more than $0.20$, so a PPO update with $\epsilon_{\mathrm{clip}} = 0.2$ would clip about one token in ten at its first step. The FP16 kernels stay within $0.05$ on almost every token.

To measure the gradient bias itself, we compute it for the gradient with respect to the output logits rather than the full parameter vector. For this gradient the score $\nabla_z \log \pi_t(y \mid x) = e_y - \pi_t(\cdot\mid x)$ has norm at most $\sqrt{2}$, so $G = \sqrt{2}$. We use a fixed advantage that gives $+1$ to the reference top-1 token, $+0.5$ to ranks 2 to 5, and $-0.5$ to all other tokens, centered under $\pi_t$ and clipped to $[-1, 1]$. We report the average norm of the per-prompt bias divided by the average norm of the per-prompt gradient. Without correction, the bias is $3.1\%$ to $3.4\%$ of the gradient for the BF16 kernels, $11.4\%$ for FP8 weights, $28.2\%$ for BF16 eager attention, and $0.4\%$ for FP16. The bound of \Cref{prop:rl-bias} is 9 to 11 times the measured bias for every kernel. Clipping the ratio as in \Cref{eq:clipped-pg} reduces the bias quickly (\Cref{fig:rl-slices}(a)). For vLLM-BF16, $c = 0.10$ lowers it from $3.08\%$ to $0.57\%$ and $c = 0.20$ lowers it to $0.05\%$.

\begin{table}[t]
\centering
\caption{Importance ratios and gradient bias when tokens are sampled from $\kinf$ and scored under \texttt{hf-fp32} with identical weights. The middle columns give the exact probability that $|w(x,y)-1|$ exceeds each threshold. The right columns give the gradient bias relative to the gradient norm, for the logit-space gradient described in the text, after clipping the ratio to $[1-c, 1+c]$; $c = 0$ is the uncorrected estimator.}
\label{tab:rl-bias}
\vspace{0.35em}
\resizebox{\linewidth}{!}{%
\begin{tabular}{lccccccccc}
\toprule
& \multicolumn{2}{c}{Sampled ratio $w(x,y)$} & \multicolumn{3}{c}{$\Pr\bigl[|w(x,y)-1| > t\bigr]$} & \multicolumn{4}{c}{Relative bias $\|\tilde{g}_c - g^\star\|_2 / \|g^\star\|_2$} \\
\cmidrule(lr){2-3} \cmidrule(lr){4-6} \cmidrule(lr){7-10}
Rollout kernel $\kinf$ & $[p_{1}, p_{99}]$ & $[\min, \max]$ & $t = 0.05$ & $t = 0.10$ & $t = 0.20$ & $c=0$ & $c=0.05$ & $c=0.10$ & $c=0.20$ \\
\midrule
\texttt{hf-fp16}        & $[0.985, 1.014]$ & $[0.953, 1.046]$ & $0.01\%$  & $0.00\%$  & $0.00\%$  & $0.42\%$  & $0.00\%$  & $0.00\%$  & $0.00\%$ \\
\texttt{vllm-fp16}      & $[0.983, 1.017]$ & $[0.960, 1.040]$ & $0.00\%$  & $0.00\%$  & $0.00\%$  & $0.41\%$  & $0.00\%$  & $0.00\%$  & $0.00\%$ \\
\texttt{vllm-bf16}      & $[0.888, 1.122]$ & $[0.719, 1.288]$ & $11.51\%$ & $2.93\%$  & $0.23\%$  & $3.08\%$  & $1.52\%$  & $0.57\%$  & $0.05\%$ \\
\texttt{vllm-bf16-b1}   & $[0.879, 1.128]$ & $[0.636, 1.807]$ & $11.96\%$ & $3.35\%$  & $0.33\%$  & $3.42\%$  & $1.82\%$  & $0.78\%$  & $0.15\%$ \\
\texttt{hf-bf16}        & $[0.882, 1.134]$ & $[0.676, 1.801]$ & $12.75\%$ & $3.91\%$  & $0.34\%$  & $3.17\%$  & $1.59\%$  & $0.64\%$  & $0.11\%$ \\
\texttt{hf-fp8-e4m3}    & $[0.610, 1.561]$ & $[0.293, 5.011]$ & $34.68\%$ & $22.39\%$ & $10.65\%$ & $11.42\%$ & $9.75\%$  & $8.18\%$  & $5.47\%$ \\
\texttt{hf-bf16-eager}  & $[0.152, 1.967]$ & $[10^{-9}, 125]$ & $46.88\%$ & $33.64\%$ & $20.48\%$ & $28.19\%$ & $25.68\%$ & $23.22\%$ & $18.62\%$ \\
\bottomrule
\end{tabular}%
}
\end{table}

\subsection{Per-slice checks catch violations that aggregate checks miss}

We define four slices of the 500 prompts. The \texttt{all} slice contains every prompt. The \texttt{high\_stakes} slice contains the 129 prompts from TruthfulQA's categories on health, medicine, nutrition, law, politics, finance, psychology, and economics. The \texttt{long\_prompt} slice contains the 79 prompts with at least 45 tokens, and the \texttt{low\_margin} slice contains the 62 prompts with $\gamma(x) < 0.50$. We apply the contract listed in \Cref{app:contract-dsl}, whose clauses are:
\begin{itemize}[leftmargin=1.8em,itemsep=0em,topsep=0.2em]
  \item \texttt{N1} (L2): on every slice, the 95th percentile of $\|\varepsilon(x)\|_\infty$ is at most $0.50$;
  \item \texttt{N-RLHF-1} (soft, L2): at most $\rho_w = 15\%$ of the prompts in any slice have $\E_{y\sim\pi_b}|\log w| > 0.05$;
  \item \texttt{N-RLHF-2} (hard, L3): on every slice, the mean of $\E_{y\sim\pi_b}[w\log w]$ is at most $0.002$;
  \item \texttt{S1} (L2): the greedy change rate over all prompts is at most $2\%$;
  \item \texttt{S3} (L3): on \texttt{high\_stakes}, the 5th percentile of the top-5 overlap is at least $0.80$;
  \item \texttt{R1} (L1): 95th-percentile latency at most $50$\,ms; \texttt{O1} (L3): replay difference at most $10^{-6}$.
\end{itemize}

\begin{figure}[t]
\centering
\includegraphics[width=\linewidth]{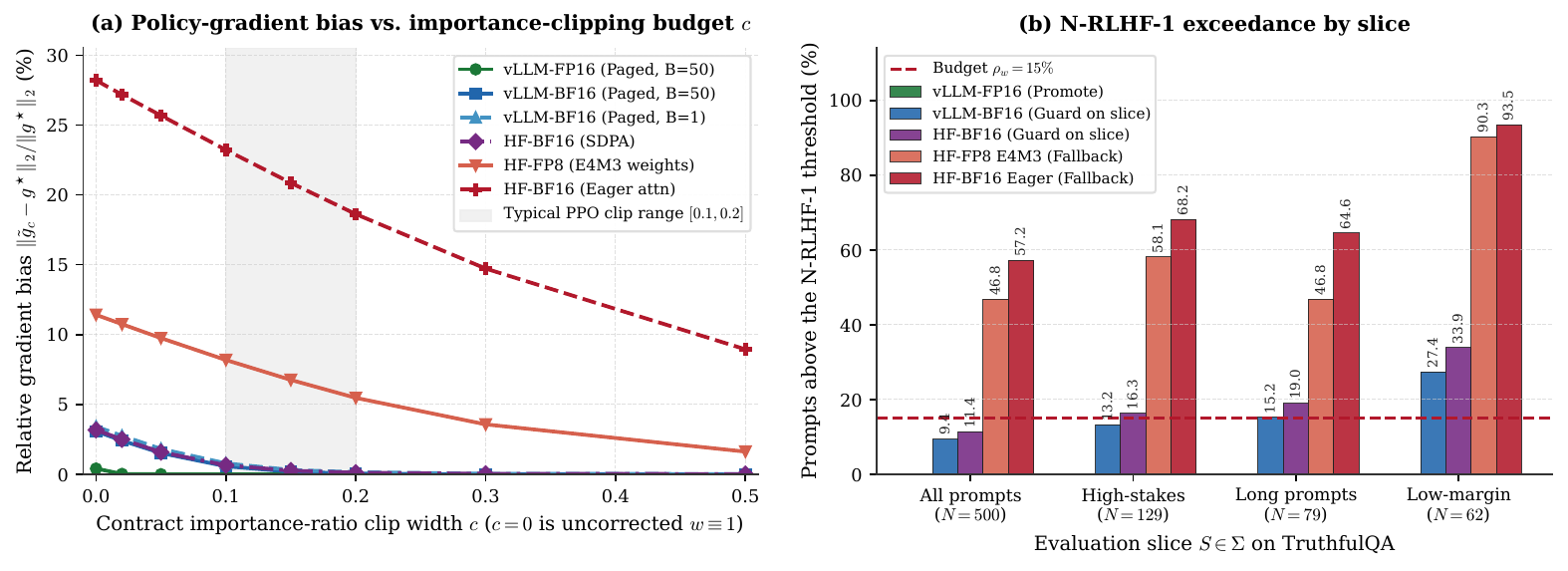}
\caption{\textbf{(a)} Gradient bias relative to the gradient norm as a function of the clip width $c$ (\Cref{prop:clipped-bias}). \textbf{(b)} Fraction of prompts in each slice that exceed the \texttt{N-RLHF-1} threshold. The dashed line is the $15\%$ budget. The BF16 kernels stay under the budget over all prompts and exceed it on the low-margin slice.}
\label{fig:rl-slices}
\end{figure}

\Cref{tab:contract-audit} and \Cref{fig:rl-slices}(b) show the results. Over all prompts, the three BF16 kernels exceed the \texttt{N-RLHF-1} threshold on $9.4\%$ to $11.4\%$ of prompts, within the $15\%$ budget, so a check on the full set alone would pass them. On the low-margin slice the rates are $27.4\%$, $32.3\%$, and $33.9\%$. HF-BF16 also exceeds the budget on the high-stakes slice ($16.3\%$) and the long-prompt slice ($19.0\%$), and vLLM-BF16 exceeds it narrowly on long prompts ($15.2\%$). The per-slice contract therefore assigns all three an L2 violation and routes them through a guarded path. vLLM-FP16 passes every clause on every slice and is promoted. HF-FP16 passes every numerical and statistical clause but fails the latency clause. Its median latency is 12\,ms, but a few slow requests raise the 95th percentile to 433\,ms, so the violation is logged at L1. FP8 weights and BF16 eager attention exceed the KL limit of \texttt{N-RLHF-2} and fall back to the reference kernel.

\begin{table}[t]
\centering
\caption{Contract results for the seven candidate kernels. Columns 2 to 5 give the fraction of prompts in each slice that exceed the \texttt{N-RLHF-1} threshold (budget $15\%$; values above the budget in bold). Column 6 gives the greedy change rate on the 62 low-margin prompts. The last two columns give the violated clauses and the action chosen by the escalation policy.}
\label{tab:contract-audit}
\vspace{0.35em}
\resizebox{\linewidth}{!}{%
\begin{tabular}{lcccccll}
\toprule
& \multicolumn{4}{c}{\texttt{N-RLHF-1} exceedance rate} & Low-margin & & \\
\cmidrule(lr){2-5}
Candidate $\kinf$ & \texttt{all} (500) & \texttt{high\_stakes} (129) & \texttt{long\_prompt} (79) & \texttt{low\_margin} (62) & greedy change & Violated clauses & Action \\
\midrule
\texttt{vllm-fp16}      & $0.0\%$  & $0.0\%$  & $0.0\%$  & $0.0\%$           & $4.8\%$  & none & promote \\
\texttt{hf-fp16}        & $0.0\%$  & $0.0\%$  & $0.0\%$  & $0.0\%$           & $4.8\%$  & \texttt{R1} & log (L1) \\
\texttt{vllm-bf16}      & $9.4\%$  & $13.2\%$ & $\mathbf{15.2\%}$ & $\mathbf{27.4\%}$ & $11.3\%$ & \texttt{N-RLHF-1} & guard (L2) \\
\texttt{vllm-bf16-b1}   & $11.2\%$ & $13.2\%$ & $10.1\%$ & $\mathbf{32.3\%}$ & $9.7\%$  & \texttt{N-RLHF-1} & guard (L2) \\
\texttt{hf-bf16}        & $11.4\%$ & $\mathbf{16.3\%}$ & $\mathbf{19.0\%}$ & $\mathbf{33.9\%}$ & $14.5\%$ & \texttt{N-RLHF-1}, \texttt{R1} & guard (L2) \\
\texttt{hf-fp8-e4m3}    & $\mathbf{46.8\%}$ & $\mathbf{58.1\%}$ & $\mathbf{46.8\%}$ & $\mathbf{90.3\%}$ & $30.6\%$ & \texttt{N1}, \texttt{N-RLHF-1}, \texttt{N-RLHF-2}, \texttt{S1} & fallback (L3) \\
\texttt{hf-bf16-eager}  & $\mathbf{57.2\%}$ & $\mathbf{68.2\%}$ & $\mathbf{64.6\%}$ & $\mathbf{93.5\%}$ & $35.5\%$ & \texttt{N1}, \texttt{N-RLHF-1}, \texttt{N-RLHF-2}, \texttt{S1}, \texttt{S3} & fallback (L3) \\
\bottomrule
\end{tabular}%
}
\end{table}

\section{Extensions}
\label{sec:future-directions}

The contracts in this paper use fixed thresholds and are checked against a fixed reference. Four extensions follow naturally.

\subsection{Thresholds that depend on the prompt}
\label{sec:fd-dynamic}

Our measurements show that greedy changes depend strongly on the reference margin $\gamma(x)$ (\Cref{fig:bounds-margins}(b)). A contract could therefore use a threshold $\tau_i(c(x))$ that depends on a summary $c(x)$ of the prompt, such as its estimated margin, its length, or its topic. One way to choose such thresholds is to trade the expected cost of violations $C(\cdot)$ against a quality constraint $Q(\cdot)$,
\begin{equation}
\tau_i^\star(c) \;=\; \argmax_{\tau} \; -\, C\bigl(\Pr[\text{violation} \mid \tau, c]\bigr) \;-\; \lambda\, Q(\tau, c),
\end{equation}
over a small, interpretable family of threshold schedules, so that the schedule itself can be versioned and reviewed with the contract.

\subsection{Training models that satisfy a contract}
\label{sec:fd-distillation}

Distillation \cite{hinton2015distilling} and preference optimization \cite{rafailov2023dpo} usually train a model under $\ktrain$ and only afterwards convert it to the serving kernel. A training objective could include smooth penalties $\phi_i$ for the contract's clauses,
\begin{equation}
\mathcal{L}_{\mathrm{total}} \;=\; \mathcal{L}_{\mathrm{task}} + \lambda_{\mathrm{KD}}\,\mathcal{L}_{\mathrm{KD}} + \sum_{i} \lambda_i\, \phi_i\bigl(m_i(\theta_s, \kinf)\bigr),
\end{equation}
for example penalizing small margins or large sensitivity to quantization, so that the trained model changes less when it is run by a low-precision kernel.

\subsection{Verifying numerical clauses}
\label{sec:fd-verification}

For numerical clauses on individual transformer blocks, sampling could be complemented by interval bound propagation and abstract interpretation \cite{katz2017reluplex,gehr2018ai2}. Translating the clauses in \Cref{app:contract-dsl} into per-block interval obligations would allow certified error bounds to be attached to a recorded combination of weights, kernel, and contract.

\subsection{Routing by margin during generation}
\label{sec:fd-routing}

Given several approved kernels with different speeds and precisions, such as vLLM in FP16, BF16, and FP8, a router could monitor the margin during generation. It could send low-margin steps or high-stakes prompts to a higher-precision kernel and serve the rest on a faster one.

\section{Related Work}
\label{sec:related}

\paragraph{Reinforcement learning from feedback and off-policy correction.}
Post-training of language models builds on TRPO \cite{schulman2015trpo}, PPO \cite{schulman2017ppo}, reinforcement learning from human feedback \cite{christiano2017deep,ouyang2022instructgpt}, DPO \cite{rafailov2023dpo}, and GRPO \cite{shao2024deepseekmath}. Off-policy reinforcement learning corrects for a difference between the behavior and target policies with importance weights \cite{precup2000eligibility,degris2012off,munos2016safe,espeholt2018impala}. In those settings the two policies usually differ because their parameters differ. In the pipelines we study the parameters are the same, and the difference comes from sampling with $\kinf$ and computing gradients with $\ktrain$. Patel \cite{patel2025rlmt} gives a step-by-step account of these algorithms for language models. Deep policy-gradient methods are known to be sensitive to implementation details \cite{henderson2018deep,engstrom2020implementation,andrychowicz2021what}, and CleanRL \cite{huang2022cleanrl} provides single-file implementations that make those details explicit. Our bounds and measurements describe one concrete mechanism by which such details can change training.

\paragraph{Serving engines and attention kernels.}
FlashAttention \cite{dao2022flashattention} and its successors \cite{dao2023flashattention2,shah2024flashattention3} reorganize the attention computation to reduce memory traffic, and vLLM \cite{kwon2023vllm} introduced PagedAttention and continuous batching. These methods compute the same function as standard attention in exact arithmetic. In floating point their outputs depend on the order of reductions, the accumulator precision, and the composition of the batch \cite{goldberg1991every,higham2002accuracy}. We measure each of these effects on \texttt{Qwen2.5-7B-Instruct} \cite{yang2024qwen25} with TruthfulQA prompts \cite{lin2022truthfulqa}.

\paragraph{Low-precision arithmetic and quantization.}
FP8 formats \cite{micikevicius2022fp8} and weight-quantization methods such as GPTQ, published as OPTQ \cite{frantar2023gptq}, and AWQ \cite{lin2023awq} usually preserve aggregate accuracy while changing individual logits. Guo et al.\ \cite{guo2017calibration} show that modern networks are often poorly calibrated, and M\"uller et al.\ \cite{mueller2019label} show that label smoothing changes calibration. Both concern the probabilities of a single model run by a single kernel. We add the observation that the effect of logit changes depends on the margin $\gamma(x)$, so that changes in the greedy token and large importance ratios concentrate on low-margin prompts.

\paragraph{Verification, hardware, and software contracts.}
Neural networks can be verified with SMT solvers \cite{katz2017reluplex} and abstract interpretation \cite{gehr2018ai2}, so far mainly for small networks. Hooker \cite{hooker2020hardware} argues that available hardware shapes which research ideas succeed. Design by contract \cite{meyer1992applying} specifies software interfaces through preconditions, postconditions, and invariants. Kernel contracts apply the same idea to the interface between a model's weights and the program that runs them.

\section{Limitations}
\label{sec:limitations}

\paragraph{Scope of the measurements.} We study one model, one type of GPU, and the first generated token of 500 short prompts with a median length of 40 tokens. Longer contexts, multi-step decoding, other model families, and other accelerators may behave differently. The FP8 results use weights rounded to E4M3 with computation in BF16, not a native FP8 matrix-multiplication kernel.

\paragraph{Per-token bounds.} \Cref{prop:rl-bias,prop:clipped-bias} bound the bias of a single token's gradient. Over a rollout of $L$ tokens, the trajectory-level importance ratio is a product of $L$ per-token ratios, and its variance can grow exponentially in $L$ \cite{precup2000eligibility,munos2016safe}. We also measured the bias only for the gradient with respect to the output logits, with a fixed advantage. We have not measured how kernel differences affect complete PPO or GRPO training runs.

\paragraph{Looseness of worst-case bounds.} \Cref{lem:logit-to-dist} is sharp for small two-token perturbations, but language-model distributions place most of their mass on a few tokens. In our measurements $\|\varepsilon(x)\|_\infty/(2T)$ is roughly an order of magnitude above the realized total-variation distance, and the bound of \Cref{prop:rl-bias} is 9 to 11 times the measured bias. Contracts should therefore limit $\Delta_{\TV}$ and $\E[w\log w]$ directly rather than rely only on limits on $\|\varepsilon\|_\infty$.

\paragraph{Cost of checking.} Numerical clauses require an extra forward pass with the reference kernel. In production this cost can be limited by checking a sample of shadow or canary traffic. In offline testing it is paid once per candidate.

\paragraph{Optimizing against the contract.} If contract clauses are used as training penalties (\Cref{sec:fd-distillation}), a model may learn to satisfy them on the checked slices while getting worse elsewhere. Keeping some slices out of training reduces this risk.

\section{Conclusion}
\label{sec:conclusion}

Training and serving pipelines run the same weights through different programs, and in finite precision these programs produce different models. We index the policy by the kernel that runs it, specify acceptable differences with kernel contracts, and prove bounds that connect logit differences to changes in the next-token distribution, in expected reward, in the greedy token, and in the policy gradient.

On \texttt{Qwen2.5-7B-Instruct}, the differences between nine Hugging Face and vLLM configurations are small on average but uneven. BF16 drifts about eight times more than FP16 from an FP32 reference, changing the engine at fixed precision drifts about as much as casting FP32 to BF16, and changing only the batch size changes the greedy token on $1.0\%$ of prompts. These changes concentrate on prompts whose two leading tokens are nearly tied, so checks on aggregate metrics pass kernels that fail on those prompts. Checking a contract on each slice makes these differences visible and ties each one to a decision about whether to promote, guard, or roll back a kernel.

\bibliographystyle{plain}
\bibliography{references}

\appendix
\crefalias{section}{appendix}
\crefalias{subsection}{appendix}
\section{A YAML Format for Kernel Contracts}
\label{app:contract-dsl}

This appendix gives a YAML format for kernel contracts and the contract used in \Cref{sec:experiments}. \Cref{app:metrics} defines the metrics it refers to.

\subsection{Format}

A contract has the following fields.
\begin{lstlisting}[language=yaml]
contract:
  id: string                 # stable contract identifier
  version: string            # semantic version (e.g., 1.0.0)
  applies_to:
    model_hashes: [string]   # checkpoint artifact hashes
    kernel_hashes: [string]  # candidate kernel build fingerprints
  slices:                    # slice definitions evaluated on D_eval
    - id: string
      filter: string         # predicate over prompt metadata / margin
  clauses:
    - id: string
      family: numerical | statistical | runtime | observability
      metric: string         # registered metric function name
      threshold: number      # bound threshold tau
      exceedance: number     # 0.0 for hard clauses, (0,1] for soft clauses
      level: L1 | L2 | L3
      slice_ids: [string]
      remediation: string    # action from escalation_policy
  escalation_policy:
    - level: L1 | L2 | L3
      action: log | guard | fallback
      target_kernel: string  # optional fallback/guard kernel identifier
\end{lstlisting}

\subsection{Contract used in the experiments}

The contract below produced the decisions in \Cref{tab:contract-audit}. The \texttt{high\_stakes} filter selects the eight TruthfulQA categories listed in \Cref{sec:experiments}.
\begin{lstlisting}[language=yaml]
contract:
  id: qwen25_gpu_v1
  version: 1.0.0
  applies_to:
    model_hashes: ["Qwen/Qwen2.5-7B-Instruct"]
    kernel_hashes: ["vllm-fp16", "vllm-bf16", "hf-bf16", "hf-fp8-e4m3"]
  slices:
    - id: all
      filter: "true"
    - id: high_stakes
      filter: "category in HIGH_STAKES_CATEGORIES"
    - id: long_prompt
      filter: "seq_len >= 45"
    - id: low_margin
      filter: "margin_ref < 0.50"
  clauses:
    - id: N1_logit_p95
      family: numerical
      metric: p95_logit_linf
      threshold: 0.50
      exceedance: 0.0
      level: L2
      slice_ids: ["all", "high_stakes", "long_prompt", "low_margin"]
      remediation: guard
    - id: N_RLHF_1_logw
      family: numerical
      metric: expected_abs_log_ratio
      threshold: 0.05
      exceedance: 0.15
      level: L2
      slice_ids: ["all", "high_stakes", "long_prompt", "low_margin"]
      remediation: guard
    - id: N_RLHF_2_wlogw
      family: numerical
      metric: mean_w_log_w
      threshold: 0.002
      exceedance: 0.0
      level: L3
      slice_ids: ["all", "high_stakes", "long_prompt", "low_margin"]
      remediation: fallback
    - id: S1_top1_flip_all
      family: statistical
      metric: top1_flip_rate
      threshold: 0.02
      exceedance: 0.0
      level: L2
      slice_ids: ["all"]
      remediation: guard
    - id: S3_top5_high_stakes
      family: statistical
      metric: p05_top5_overlap
      threshold: 0.80
      exceedance: 0.0
      level: L3
      slice_ids: ["high_stakes"]
      remediation: fallback
    - id: R1_latency_p95_ms
      family: runtime
      metric: p95_latency_ms
      threshold: 50.0
      exceedance: 0.0
      level: L1
      slice_ids: ["all"]
      remediation: log
    - id: O1_deterministic_replay
      family: observability
      metric: replay_max_abs_diff
      threshold: 1.0e-6
      exceedance: 0.0
      level: L3
      slice_ids: ["all"]
      remediation: fallback
  escalation_policy:
    - level: L1
      action: log
    - level: L2
      action: guard
      target_kernel: vllm-fp16
    - level: L3
      action: fallback
      target_kernel: hf-fp32
\end{lstlisting}

\section{Proofs}
\label{app:proofs}

\subsection{Proof of \texorpdfstring{\Cref{lem:logit-to-dist}}{Lemma 1}}
\label{app:proof-logit-to-dist}

Write $p = \mathrm{softmax}(z/T)$ and $q = \mathrm{softmax}((z + \varepsilon)/T)$. Let $\delta = \varepsilon / T$, $m = \min_{v} \delta_v$, $M = \max_{v} \delta_v$, and $S = M - m = \spanop(\varepsilon)/T \le 2\,\|\varepsilon\|_\infty / T$.

\paragraph{Ratio bound (\Cref{eq:lem-ratio}).}
For every $v$,
\begin{equation}
\frac{q(v)}{p(v)} \;=\; \frac{e^{\delta_v}}{\sum_{u} p(u)\,e^{\delta_u}}.
\end{equation}
The numerator lies in $[e^m, e^M]$, and the denominator is an average of numbers in $[e^m, e^M]$, so it lies in the same interval. The ratio therefore lies in $[e^{-S}, e^{S}]$, and so does its reciprocal $p(v)/q(v)$.

\paragraph{Linear total-variation bound (\Cref{eq:lem-tv}).}
For $t \in [0, 1]$ let $p_t = \mathrm{softmax}(z/T + t\delta)$, so that $p_0 = p$ and $p_1 = q$. Then
\begin{equation}
\frac{d}{dt} p_t(v) \;=\; p_t(v)\,\bigl(\delta_v - \mu_t\bigr), \qquad \mu_t = \E_{u \sim p_t}[\delta_u].
\end{equation}
Among distributions on $[m, M]$ with mean $\mu_t$, the mean absolute deviation $\E|\delta - \mu_t|$ is largest for the distribution supported on the two endpoints, because $x \mapsto |x - \mu_t|$ is convex. That largest value is $2(M - \mu_t)(\mu_t - m)/(M - m) \le (M - m)/2$. Hence $\sum_v |dp_t(v)/dt| \le S/2$ for every $t$, and
\begin{equation}
\Delta_{\TV}(p, q) \;=\; \tfrac{1}{2}\sum_{v} \Bigl|\int_0^1 \frac{d}{dt} p_t(v)\,dt\Bigr| \;\le\; \tfrac{1}{2}\int_0^1 \sum_{v} \Bigl|\frac{d}{dt} p_t(v)\Bigr|\,dt \;\le\; \frac{S}{4} \;=\; \frac{\spanop(\varepsilon)}{4T} \;\le\; \frac{\lVert \varepsilon \rVert_\infty}{2T}.
\end{equation}

\paragraph{Hyperbolic-tangent bound (\Cref{eq:lem-tv}).}
Let $A = \{v : q(v) \ge p(v)\}$, $P = p(A)$, and $Q = q(A)$, so that $\Delta_{\TV}(p, q) = Q - P$. Since $q(A) \propto \sum_{v \in A} p(v) e^{\delta_v} \le e^{M} P$ and $q(A^c) \propto \sum_{v \notin A} p(v) e^{\delta_v} \ge e^{m}(1 - P)$ with the same normalizing constant,
\begin{equation}
\frac{Q}{1 - Q} \;\le\; e^{S}\,\frac{P}{1 - P}, \qquad\text{so}\qquad Q - P \;\le\; \frac{e^{S} P}{e^{S} P + 1 - P} - P.
\end{equation}
The right-hand side is maximized over $P \in [0, 1]$ at $P = 1/(1 + e^{S/2})$, where it equals $(e^{S/2} - 1)/(e^{S/2} + 1) = \tanh(S/4)$. Since $\tanh(u) \le u$ for $u \ge 0$, this bound is never weaker than the linear one. It is attained by the two-token distribution described after \Cref{lem:logit-to-dist}.

\paragraph{KL bound (\Cref{eq:lem-kl}).}
Let $\psi(t) = \log \sum_{v} p(v)\,e^{t\delta_v}$. Then $\psi(0) = 0$, $\psi'(t) = \mu_t$, and $\psi''(t) = \mathrm{Var}_{p_t}(\delta)$. Since $\log(p(v)/q(v)) = \psi(1) - \delta_v$,
\begin{equation}
\KL(p \,\Vert\, q) \;=\; \psi(1) - \psi'(0) \;=\; \int_0^1 (1-t)\,\psi''(t)\,dt \;=\; \int_0^1 (1-t)\,\mathrm{Var}_{p_t}(\delta)\,dt.
\end{equation}
By Popoviciu's inequality, $\mathrm{Var}_{p_t}(\delta) \le (M-m)^2/4 = S^2/4$. Since $\int_0^1 (1-t)\,dt = 1/2$,
\begin{equation}
\KL(p \,\Vert\, q) \;\le\; \frac{S^2}{8} \;=\; \frac{\spanop(\varepsilon)^2}{8T^2} \;\le\; \frac{\lVert \varepsilon \rVert_\infty^2}{2T^2}. \qquad\square
\end{equation}

\subsection{Proof of \texorpdfstring{\Cref{lem:margin-flip}}{Lemma 2}}
\label{app:proof-margin-flip}

Let $v^\star$ be the reference greedy token. By the definition of $\gamma(x)$, every other token $u$ satisfies $z^{\ktrain}_\theta(x)_{v^\star} - z^{\ktrain}_\theta(x)_u \ge \gamma(x)$. Under $\kinf$,
\begin{align}
z^{\kinf}_\theta(x)_{v^\star} - z^{\kinf}_\theta(x)_u
&\;=\; \bigl(z^{\ktrain}_\theta(x)_{v^\star} - z^{\ktrain}_\theta(x)_u\bigr) + \bigl(\varepsilon_{v^\star}(x) - \varepsilon_u(x)\bigr) \nonumber\\
&\;\ge\; \gamma(x) - \spanop\bigl(\varepsilon(x)\bigr).
\end{align}
If $\spanop(\varepsilon(x)) < \gamma(x)$, the right-hand side is positive for every $u \neq v^\star$, so $v^\star$ is also the greedy token under $\kinf$. Since $\spanop(\varepsilon(x)) \le 2\|\varepsilon(x)\|_\infty$, the condition $\|\varepsilon(x)\|_\infty < \gamma(x)/2$ is sufficient. \hfill$\square$

\subsection{Proof of \texorpdfstring{\Cref{prop:reward-bound}}{Proposition 1}}
\label{app:proof-reward-bound}

Since $|r(x,y)| \le R$,
\begin{align}
\bigl|\Delta_r(x)\bigr|
&\;=\; \Bigl|\sum_{y} \bigl(\polktrain_\theta(y\mid x) - \polkinf_\theta(y\mid x)\bigr)\,r(x,y)\Bigr| \nonumber\\
&\;\le\; R \sum_{y} \bigl|\polktrain_\theta(y\mid x) - \polkinf_\theta(y\mid x)\bigr| \;=\; 2R\,\Delta_{\TV}(x).
\end{align}
The second inequality in \Cref{eq:prop-reward} follows from $\Delta_{\TV}(x) \le \|\varepsilon(x)\|_\infty / (2T)$ (\Cref{lem:logit-to-dist}). \hfill$\square$

\subsection{Proof of \texorpdfstring{\Cref{prop:rl-bias}}{Proposition 2}}
\label{app:proof-rl-bias}

By (A1), writing the expectations over $y \sim \pi_b(\cdot\mid x)$ as sums gives
\begin{align}
g^\star(\theta) &\;=\; \E_{x \sim D}\left[ \sum_{y} \pi_t(y\mid x)\, \nabla_\theta \log \pi_t(y\mid x) \, A(x,y) \right],\\
\hat g(\theta)  &\;=\; \E_{x \sim D}\left[ \sum_{y} \pi_b(y\mid x)\, \nabla_\theta \log \pi_t(y\mid x) \, A(x,y) \right].
\end{align}
Subtracting, and using the triangle inequality with (A2) and (A3),
\begin{align}
\lVert \hat g(\theta) - g^\star(\theta) \rVert_2
&\;\le\; \E_{x \sim D}\left[ \sum_{y} \bigl|\pi_b(y\mid x) - \pi_t(y\mid x)\bigr| \, \lVert \nabla_\theta \log \pi_t(y\mid x) \rVert_2 \, \bigl|A(x,y)\bigr| \right] \nonumber\\
&\;\le\; A_{\max}\, G \cdot \E_{x \sim D}\bigl[ \lVert \pi_b(\cdot \mid x) - \pi_t(\cdot\mid x) \rVert_1 \bigr] \;=\; 2 A_{\max}\, G \cdot \E_x\bigl[\Delta_{\TV}(x)\bigr].
\end{align}
The last inequality in \Cref{eq:prop-rl-bias} follows from \Cref{lem:logit-to-dist}. For \Cref{eq:prop-rl-pinsker}, Pinsker's inequality gives $\Delta_{\TV}(x) \le \sqrt{\Delta_{\KL}(x)/2}$, \Cref{eq:wlogw-kl} gives $\Delta_{\KL}(x) = \E_{y\sim\pi_b}[w\log w]$, and Jensen's inequality moves the expectation over $x$ inside the square root. \hfill$\square$

\subsection{Proof of \texorpdfstring{\Cref{prop:clipped-bias}}{Proposition 3}}
\label{app:proof-clipped-bias}

Because $\pi_b(y\mid x) > 0$,
\begin{align}
\tilde{\pi}_c(y\mid x) &\;=\; \pi_b(y\mid x)\,\mathrm{clip}\bigl(w(x,y),\, 1-c,\, 1+c\bigr) \nonumber\\
&\;=\; \mathrm{clip}\bigl(\pi_t(y\mid x),\, (1-c)\pi_b(y\mid x),\, (1+c)\pi_b(y\mid x)\bigr),
\end{align}
and therefore
\begin{equation}
\bigl|\tilde{\pi}_c(y\mid x) - \pi_t(y\mid x)\bigr| \;=\; \Bigl(\bigl|\pi_t(y\mid x) - \pi_b(y\mid x)\bigr| - c\,\pi_b(y\mid x)\Bigr)_{+}.
\end{equation}
The bias is $\tilde{g}_c(\theta) - g^\star(\theta) = \E_x\bigl[\sum_y (\tilde{\pi}_c(y\mid x) - \pi_t(y\mid x))\,\nabla_\theta \log \pi_t(y\mid x)\,A(x,y)\bigr]$, and the argument of \Cref{app:proof-rl-bias} gives \Cref{eq:clipped-bias-bound}. Each summand $(|\pi_t - \pi_b| - c\,\pi_b)_+$ is non-increasing and convex in $c$, equals $|\pi_t - \pi_b|$ at $c = 0$, and is zero once $c \ge |w(x,y) - 1|$. \hfill$\square$

\section{Metric Definitions}
\label{app:metrics}

\Cref{tab:metric-defs} defines the metrics named in the contract of \Cref{app:contract-dsl}. A clause passes on a slice if its metric is within the threshold. For a soft clause on a per-prompt metric, the clause passes if the fraction of prompts in the slice that exceed the threshold is at most the exceedance budget. The escalation policy then applies the action of the most severe failed clause across all slices, or promotes the candidate if every clause passes.

\begin{table}[ht]
\centering
\caption{Metrics used in the contract of \Cref{app:contract-dsl}. Slice statistics are computed over the prompts $x$ in a slice $S$.}
\label{tab:metric-defs}
\vspace{0.35em}
\small
\begin{tabular}{lp{0.62\linewidth}}
\toprule
Metric & Definition \\
\midrule
\texttt{p95\_logit\_linf} & 95th percentile over $x \in S$ of $\|\varepsilon(x)\|_\infty$. \\
\texttt{expected\_abs\_log\_ratio} & Per-prompt value $\E_{y \sim \polkinf_\theta(\cdot\mid x)} |\log w(x,y)|$, computed over the full vocabulary. \\
\texttt{mean\_w\_log\_w} & Mean over $x \in S$ of $\E_{y \sim \polkinf_\theta(\cdot\mid x)}[w(x,y)\log w(x,y)] = \Delta_{\KL}(x)$. \\
\texttt{top1\_flip\_rate} & Fraction of $x \in S$ whose greedy token differs between the two kernels, $1 - \E_{x \in S}[A_{\mathrm{top1}}(x)]$. \\
\texttt{p05\_top5\_overlap} & 5th percentile over $x \in S$ of $A_{\mathrm{top5}}(x)$. \\
\texttt{p95\_latency\_ms} & 95th percentile of per-request latency in milliseconds. \\
\texttt{replay\_max\_abs\_diff} & Largest absolute difference between the logits of two runs of the same prompts on the same kernel. \\
\bottomrule
\end{tabular}
\end{table}

\end{document}